\pdfoutput=1

\documentclass[11pt]{article}

\usepackage[preprint]{acl}

\usepackage{times}
\usepackage{latexsym}
\usepackage{amsmath}
\usepackage{amsfonts}
\usepackage{algorithm}
\usepackage{algorithmic}
\usepackage{multicol}
\usepackage{multirow}
\usepackage{adjustbox}
\usepackage{booktabs}

\usepackage[T1]{fontenc}

\usepackage[utf8]{inputenc}

\usepackage{microtype}

\usepackage{inconsolata}

\usepackage{graphicx}

\title{Reasoning-Enhanced Self-Training for Long-Form \\ Personalized Text Generation}

\author{Alireza Salemi\textsuperscript{1}\thanks{Work done as a student researcher at Google DeepMind.}, Cheng Li\textsuperscript{2}, Mingyang Zhang\textsuperscript{2}, Qiaozhu Mei\textsuperscript{3}\thanks{Work done as a vsiting researcher at Google DeepMind.}, Weize Kong\textsuperscript{2} \\ \textbf{Tao Chen\textsuperscript{2}, Zhuowan Li\textsuperscript{2}, Michael Bendersky\textsuperscript{2},  Hamed Zamani\textsuperscript{1}}\\\\
\textsuperscript{1}University of Massachusetts Amherst \quad \textsuperscript{2}Google DeepMind \quad
\textsuperscript{3}University of Michigan \\
\textsuperscript{1}\texttt{\{asalemi, zamani\}@cs.umass.edu} \\ 
\textsuperscript{2}\texttt{\{chgli, mingyang, weize, taochen, zhuowan, bemike\}@google.com}\\
\textsuperscript{3}\texttt{qmei@umich.edu}
}

\begin{document}
\maketitle
\begin{abstract}

Personalized text generation requires a unique ability of large language models (LLMs) to learn from context that they often do not encounter during their standard training. One way to encourage LLMs to better use personalized context for generating outputs that better align with the user's expectations is to instruct them to reason over the user’s past preferences, background knowledge, or writing style. To achieve this, we propose Reasoning-Enhanced Self-Training for Personalized Text Generation (REST-PG), a framework that trains LLMs to reason over personal data during response generation. REST-PG first generates reasoning paths to train the LLM's reasoning abilities and then employs Expectation-Maximization Reinforced Self-Training to iteratively train the LLM based on its own high-reward outputs. We evaluate REST-PG on the LongLaMP benchmark, consisting of four diverse personalized long-form text generation tasks. Our experiments demonstrate that REST-PG achieves significant improvements over state-of-the-art baselines, with an average relative performance gain of 14.5\% on the benchmark.
\end{abstract}

\section{Introduction}

Personalizing large language models (LLMs) emerges as a critical topic in natural language processing \cite{lamp,longlamp}, due to its wide-ranging applications in recommender systems \cite{llm-rs-tutorial, chen2023surveylargelanguagemodels}, virtual assistants \cite{li2024helloagainllmpoweredpersonalized, info:doi/10.2196/15360}, and content generation \cite{alhafni-etal-2024-personalized}. The importance of personalization in such systems stems from the fact that they provide targeted content to their users, which enhances user satisfaction, improves engagement, and increases efficiency.

Augmenting the input context of the LLMs with retrieved personalized context alongside the user prompt has proven effective in tailoring responses to individual users \cite{lamp, rspg}. However, defining the notion of relevance, a prerequisite for retrieving personalized context, is challenging \cite{rspg}. In personalization, a part of the user's context that appears not directly ``relevant'' to the prompt might be more useful (than a directly relevant one) if it better reflects the user's implicit preferences. For example, a sentence like ``\textit{I have two children of age 3 and 4...}'' in the user context does not seem directly relevant to the prompt ``\textit{Give some suggestions about brands of room heaters.}'' However, this knowledge indicates that the user could be concerned about safety for children and therefore would expect the model to consider this in its response. Establishing such an ``implicit'' relevance requires reasoning beyond the words or semantics of the user context, just like the user themselves does. We argue that an approach for encouraging an  LLM to better use personalized context is also asking it to reason over it prior to generating the final response. For instance, the model may summarize the user's writing style, interests, background knowledge, and preferences before actually responding to the user prompt. However, it is often infeasible or costly to obtain sufficient human reasoning paths to train an LLM for personalized reasoning. 

This paper addresses these challenges by introducing \underline{R}easoning-\underline{E}nhanced \underline{S}elf-\underline{T}raining for \underline{P}ersonalized Text \underline{G}eneration (REST-PG), a multi-stage framework designed to teach LLMs reasoning over personalized context through reinforced self-training. As an alternative to human reasoning paths, REST-PG uses an LLM to generate the reasoning steps considering the input, expected output, and personalized context. These generated reasoning paths are then used to train the LLM, through supervised fine-tuning, to produce both the reasoning steps and the final response in a single inference path. Nevertheless, we find that supervised fine-tuning on generated reasoning data alone is insufficient for training the LLMs to produce both the reasoning path and final response, and exploring diverse reasoning paths plays a key role in obtaining effective personalized outputs; we observe a drop in performance compared to an LLM without reasoning. This suggests that the reasoning paths generated by the fine-tuned LLMs may not yet align well with the user's preferences. 
To address this, we employ Expectation-Maximization Reinforced Self-Training, which optimizes the model to generate reasoning paths that yield better aligned responses—-i.e., responses that achieve higher rewards. In an Expectation (E) step, the LLM generates different reasoning paths and responses for each input. In a Maximization (M) step, the reasoning paths that result in high-reward responses---those with high similarity to the expected output for the user---are then used to train the LLM in subsequent iterations. Through iterative process of expectation maximization, the LLM learns to generate reasoning steps and responses that are more aligned with the user's preferences.

We perform our experiments on the Long-form Language Model Personalization (LongLaMP) benchmark \cite{longlamp}, comprising four diverse long-form personalized text generation tasks. Experiments on this benchmark show that REST-PG on average significantly outperforms all state-of-the-art baseline models across all tasks of the LongLaMP benchmark. Specifically, REST-PG improves performance by up to 14.5\% compared to supervised fine-tuning (SFT) and by 6.5\% compared to self-training without reasoning enhancement. Additionally, our extensive ablation study provides valuable insights into various components of the proposed method about self-training and reasoning in personalizing LLMs.

\section{Problem Formulation}

This paper addresses personalized text generation, a task that uses user-specific information to tailor responses to individual users. A general LLM $M_\theta$ generates a piece of text in response to an input prompt $x$ from a user $u$, denoted as $\hat{y} = M_\theta(x)$. To personalize an LLM for the user $u$, we assume each prompt $x$ from the user, with the expected output $y$, is accompanied by the user profile $P_u = \{d_{(u, i)}\}_{i = 1}^{|P_u|}$, consisting of unstructured information pieces about the user $u$. Accordingly, we assume access to training and evaluation data $D = \{(x_i, y_i, P_i)\}_{i = 1} ^ {|D|}$ in the above format. Our primary objective is to utilize the personalized information from user profile with LLM $M_\theta$ to generate a response $\hat{y} = M_\theta(x, P_u)$ that maximizes the reward $r = \mathcal{R}(x, y, \hat{y})$ generated from a reward function $\mathcal{R}$ given the input prompt $x$, the expected output $y$ by the user $u$, and the actually generated response $\hat{y}$. The primary objective of the reward model is to evaluate the similarity between the generated output and the expected personalized output.

\section{REST-PG}

\begin{figure}
    \centering
    \includegraphics[width=\linewidth]{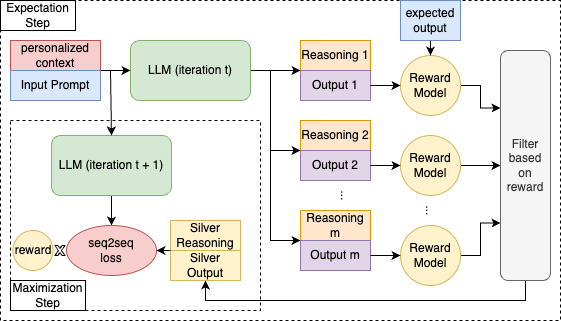}
    \caption{The overview of training pipeline of Reasoning-Enhanced Self-Training for Personalized Text Generation (REST-PG).}
    \label{fig:method}
    \vspace{-0.5cm}
\end{figure}

LLMs have proven effective in learning from their context \cite{wei2022emergent, NEURIPS2020_1457c0d6}, making the augmentation of their input with personalized context an effective strategy for personalizing their responses \cite{lamp, peft-rag-personalization}. However, learning to personalize from context requires a specialized form of context-based learning, as it involves not only understanding task-relevant information but also inferring user-specific preferences. For instance, a sentence in the personalized context that is seemingly irrelevant to the user prompt could indicate implicit preference, like mentioning children could imply prioritizing safety.
Teaching LLMs to recognize this nuanced notion of relevance is crucial for improving personalized text generation. One approach to do this is to instruct LLMs to reason over the personalized context by generating a summary of the user's preferences before responding to the prompt. However, collecting training data for this is challenging, as human annotations are costly and often fail to accurately capture the nuances of individual user preferences. To address this, LLMs can be used to generate reasoning paths based on the personalized context, input prompt, and expected output to creating reasoning training data that guides the model from the input to the output without relying on human annotation. This data can be used to train LLMs to do reasoning during personalized response generation.

While this approach seems effective, the generated reasoning paths are based on the model's implicit understanding of user preferences, which may not always align with the actual preferences. To address this, the LLM can be trained to improve this alignment by optimizing a reward function that evaluates the user's satisfaction by comparing the generated output with the expected output for that user. This alignment pushes the model toward generating reasoning paths that lead to responses more consistent with the user's preferences. This paper focuses on training the LLM to reason over personalized contexts and to generate personalized outputs  in a single inference path. Figure \ref{fig:method} provides an overview of the optimization approach used in this paper. We employ Expectation-Maximization Reinforced Self-Training \cite{rest-em} as a preference alignment algorithm to self-train the LLM, enhancing its ability to generate reasoning paths that lead to more effective personalized outputs, according to a reward model that considers the expected personalized output. This enables the model to better leverage user-specific context with improved reasoning ability, ultimately improving the quality of the generated personalized responses.

\begin{algorithm*}
    \caption{\underline{R}easoning-\underline{E}nhanced \underline{S}elf-\underline{T}raining for \underline{P}ersonalized Text \underline{G}eneration (REST-PG).}
    \label{alg:main}
    \begin{algorithmic}[1]
        \STATE \textbf{Input:} training dataset $D$, training LLM $M_\theta$, data generation LLM for preference summarization
        \STATE \textcolor{blue}{// generating the reasoning data}
        \STATE $D_{\text{reasoning}} = \{(x,\text{concat}(reasoning, y), P) | (x, y, P) \sim D : reasoning = \text{LLM}(x, y, P)\}$ 
        \STATE \textcolor{blue}{// SFT on the reasoning dataset}
        \STATE ${\theta^1} = {\mathrm{arg\,max}}_\theta \mathop{\mathbb{E}}_{(x, y, P) \sim D_{\text{reasoning}}}\left[ \log p_{\theta}(y| x;P) \right]$
        \STATE \textcolor{blue}{// training the model for T iterations}
        \FOR{$t = 1$ to $T$}
            \STATE \textcolor{blue}{//  Expectation step: generating different reasoning paths and outputs to be rewarded}
            \STATE $D_{t} = \{(x, y, P, \hat{y}_j)| (x, y, P) \sim D, \hat{y}_j \sim M_{\theta^t}(x, P):  \mathcal{R}(x, y, \hat{y}_j) \geq \tau\}$ 
            \STATE \textcolor{blue}{// Maximization step: maximizing the probability of the outputs with high reward}
            \STATE $\theta^{t+1} = {\mathrm{arg\,max}}_\theta \mathop{\mathbb{E}}_{(x, y, \hat{y}, P) \sim D_{t}}\left[ \mathcal{R}(x, y, \hat{y})\log p_{\theta}(\hat{y}| x;P) \right]$ 
        \ENDFOR
    \end{algorithmic}
\end{algorithm*}

\subsection{Enhancing Personalization by Reasoning}
\label{sec:reasoning-data-gen}

Current state-of-the-art methods for personalizing LLMs augment the input with a personalized context (often retrieved from a personal corpus) \cite{lamp,rspg,longlamp}. We argue that effectively utilizing personalized context necessitates a specialized form of context-based learning, as it requires understanding both task-relevant information and user-specific preferences—an aspect that LLMs are rarely exposed to during standard training. One way to encourage LLMs to better utilize personalized context is to instruct them to focus on user-specific elements such as preferences, interests, background knowledge, and writing style that are present in the personalized context. Incorporating these attributes from the personalized context enables the model to generate more aligned, user-specific responses. These attributes can be inferred by the LLM through reasoning over the personalized context, enabling it to interpret the user's preferences, interests, knowledge, and writing style before generating the final personalized response to the user's prompt. This reasoning step helps the model produce more accurate and personalized outputs.

To generate the necessary data for training the LLM to perform such reasoning steps, we introduce a semi-supervised data generation method tailored for this purpose. In this method, for a given input $x$ for user $u$, the user profile $P_u$, and the expected output $y$, we use an LLM\footnote{We utilize Gemma 7B \cite{gemma} as the LLM to generate preliminary reasoning data.} to generate a summary of user's preferences, interests, background knowledge, and writing style features tailored to the given input and corresponding output from the user context. The detailed prompt is presented in Figure~\ref{fig:summary_gen_prompt} in Appendix~\ref{app:prompts}. This prompt encourages the model to take into account both the expected output and the input, and based on this, generate its interpretation of the user's interests, preferences, and familiarity with various topics from the personalized context as a reasoning path. Additionally, the approach guides the model to infer patterns in the user's preferences across different topics. For instance, if the user writes about a specific topic in a particular style, the model can generalize this pattern, assuming the user might adopt a similar style for other topics as well. Figures \ref{fig:generated-summary-abstract} and \ref{fig:generated-summary-review} in Appendix \ref{app:case-study} present some examples of the generated reasoning paths. These figures illustrate how the model reasons over the personalized context by analyzing the key aspects of user's preferences.

Finally, to train the LLM to reason over personalized context during output generation, the generated reasoning over personalized context is combined with the expected output using a predefined template, as shown in Figure~\ref{fig:cot-input-output-template} in Appendix \ref{app:prompts}. This template allows us to train the model to generate this combined output given an input from a specific user accompanied by its personalized context. The model is first asked to generate a summary of the user's preferences and writing style features based on the input and personalized context then generates a response to the input. Here, the combined generated reasoning and expected output are used as the new expected output for the corresponding input in the template. Indeed, the model's task is to generate both the reasoning path, based on the personalized context, and the final response in a single inference pass. This structured approach helps the LLM learn to incorporate reasoning over the personalized context as the steps toward generating the final response to the input.

\subsection{Reasoning-Enhanced Self-Training}

While we can train the model using SFT on the generated reasoning data from Section \ref{sec:reasoning-data-gen} so that it reasons towards generating personalized responses, the reasoning itself is derived from the LLM’s interpretation of the user profile, input prompt, and expected output. This reliance on the LLM's implicit understanding introduces potential limitations, as the reasoning path may not fully align with the user's preferences. Moreover, there is no guarantee that the generated reasoning path can consistently improve the final output. There may exist alternative reasoning paths that lead to more effective personalized responses, which are not captured by the initially generated reasoning paths for SFT. 

A solution to address this is to employ RL, which allows the model to explore the trajectory space (i.e., reasoning paths) to identify those that lead to personalized outputs with higher rewards. By leveraging exploration, the model can discover reasoning paths that yield higher rewards, corresponding to more desirable personalized outputs. Specifically, we employ Expectation-Maximization Reinforced Self-Training \cite{rest-em} as an offline RL algorithm to encourage the model to discover reasoning paths that lead to higher rewards. The algorithm used for this purpose is detailed in Algorithm \ref{alg:main}. After performing SFT on the data generated in Section \ref{sec:reasoning-data-gen}, we iteratively alternate between the following steps:

\paragraph{Expectation Step:} 
In this step, the optimized parameter set from the previous iteration (i.e., $\theta^t$) is used to collect new trajectories for training the model for the next iteration (i.e., $\theta^{t+1}$). Specifically, for each input $x \in D$, the LLM $M_{\theta^t}$ is employed to generate $m$ outputs using a decoding temperature $\gamma$. The temperature $\gamma$ controls the amount of randomness in the generated outputs, which indicates the freedom of the model in the exploration phase of the reinforcement learning algorithm. The generated outputs are then evaluated using the reward model, denoted as $\mathcal{R}(x, y, \hat{y}_j)$. The reward model focuses solely on the final output generated by the model, disregarding the reasoning path itself, and assigns a score to each output. Thus, the reward model only considers the similarity between the generated response and expected output to score the reasoning paths. Finally, the outputs that achieve a reward of $\tau$ or higher are considered high quality outputs and are included in the next round of training data, where they act as the expected output for the corresponding inputs. To prevent the model from overfitting on easy examples, we limit the number of outputs retained per input to a maximum of 10 to ensures diverse outputs and avoid overfitting to simpler cases.

\paragraph{Maximization Step:} 

This step uses the dataset generated from the expectation step to optimize the model. In this phase, the outputs that received high rewards are used as the expected outputs for their corresponding inputs. Furthermore, the weight of each output is adjusted according to the reward it receives, as detailed in Algorithm \ref{alg:main} (line 11). Indeed, instead of maximization, a SFT sequence-to-sequence loss \cite{seq2seq} can be minimized to train the LLM,\footnote{Minimizing seq2seq loss corresponds to maximizing likelihood of generating the ground-truth sequence.} with the loss being adjusted based on the amount of reward each output receives. The underlying idea is that samples resulting in higher rewards should have a larger impact on the loss. This approach ensures that the model learn more from high-reward examples, helping it generate high-quality, personalized responses.

\section{Experiments}
\label{sec:experiments}

\begin{table*}
    \centering
    \adjustbox{max width=\textwidth}{
    \begin{tabular}{ll|c|c|c|c||c}
        \toprule
        & \multirow{3}{*}{\textbf{Model}} & \textbf{LongLaMP-1:} & \textbf{LongLaMP-2:} & \textbf{LongLaMP-3:} & \textbf{LongLaMP-4:} & \multirow{2}{*}{\textbf{Average}} \\
        & & \textbf{Personalized} & \textbf{Personalized} & \textbf{Personalized} & \textbf{Personalized} & \\
        & & \textbf{Email Completion} & \textbf{Abstract Generation} & \textbf{Review Writing} & \textbf{Topic Writing} & \textbf{(macro)} \\
         \midrule
        1 & SFT  & 0.2974 & 0.4135$^2$ & 0.6525 & 0.2270 & 0.3976 \\
        
        2 & SFT w/ Reasoning-Enhancement & 0.2834 & 0.3829 & 0.6773$^1$ & 0.2184 & 0.3905 \\

        3 & ReST-EM  & 0.3032 & 0.4549$^{12}$ & 0.6656 & 0.2859$^{12}$ & 0.4274$^{12}$ \\
        \midrule
        
        & \textbf{REST-PG} & \textbf{0.3059} & \textbf{0.4845$^{123}$} & \textbf{0.7077$^{123}$} & \textbf{0.3238$^{123}$} & \textbf{0.4554$^{123}$} \\
        \bottomrule
    \end{tabular}}
    \caption{The performance of all methods on the test sets of the LongLaMP benchmark. The superscripts 1, 2, and 3 denote statistically significant improvements compared to the model in the corresponding row using the two-tailed paired t-test ($p < 0.05$). The results on the validation sets are reported in Table~\ref{tab:main-results-dev} in Appendix \ref{app:results-dev}.}
    \label{tab:main-results}
    \vspace{-0.5cm}
\end{table*}

\subsection{Experimental Setup}

\paragraph{Datasets.}

We adopt the LongLaMP benchmark \cite{longlamp} to conduct our experiments, which consists of four personalized long-form text generation tasks: (1) Personalized Email Completion, (2) Personalized Abstract Generation, (3) Personalized Review Writing, and (4) Personalized Topic Writing. Each example in this dataset represents a separate user, including an input prompt, an expected output, and a user profile containing information about the user (i.e., documents written by the user over time). This setup allows us to evaluate the effectiveness of our approach in generating personalized responses across diverse tasks. More details about the datasets in the LongLaMP benchmark are provided in Appendix \ref{app:setup}.

\paragraph{Reward Modeling \& Evaluation.}

While the LongLaMP benchmark uses ROUGE metrics \cite{rouge} for evaluating long-form generated text, previous research shows that term-matching metrics like ROUGE often struggle to capture nuanced text similarities \cite{bert-score}, particularly in long-form text generation \cite{koh-etal-2022-far, krishna-etal-2021-hurdles}. Following recent text generation evaluation approaches \cite{gao2024llmbasednlgevaluationcurrent, geval}, we use Gemma 7B \cite{gemma} as the evaluator. We provide the evaluator LLM with the input prompt, the generated output, and the expected output, along with a prompt that explains the evaluation criteria, as shown in Figure \ref{fig:eval-prompt} in Appendix \ref{app:human-eval}. The score for the generated output is computed based on the probability that the LLM assigns to each criterion. Specifically, the final score is the score of the criterion that has the maximum probability of being generated given the evaluation prompt. Finally, we normalize this score in range of 0 and 1 by dividing it by 10. The details of the reward model are explained in Appendix \ref{app:human-eval}.

\paragraph{Training \& Inference Setting.}

We use Gemma 2B \cite{gemma} as the personalized generator LLM. Given that user profiles can contain numerous items, making it impractical to use all of them, we utilize RAG to integrate personalized context \cite{lamp}. We employ the prompt illustrated in Figure \ref{fig:cot-input-output-template} in Appendix \ref{app:prompts}, where we retrieve \(k = 5\) items from the user profile using Contriever \cite{contriever}, based on their similarity of the items to the input. Since LLMs have been shown to effectively handle multiple tasks concurrently, we train a single model on all datasets. Following \citet{rest-em}, the models are trained for $T = 3$ iterations, generating $m = 32$ outputs for each input during the expectation step with temperature $\gamma = 0.7$ using Nucleus Sampling \cite{Holtzman2020The}, unless otherwise specified. We set the output selection threshold $\tau=1.0$. At each iteration, we start from a new untrained checkpoint unless otherwise noted. For inference, temperature $\gamma = 0.1$ is used. The details are provided in Appendix \ref{app:setup}.

\subsection{Main Findings}

\begin{figure*}
    \centering
    \includegraphics[width=\textwidth]{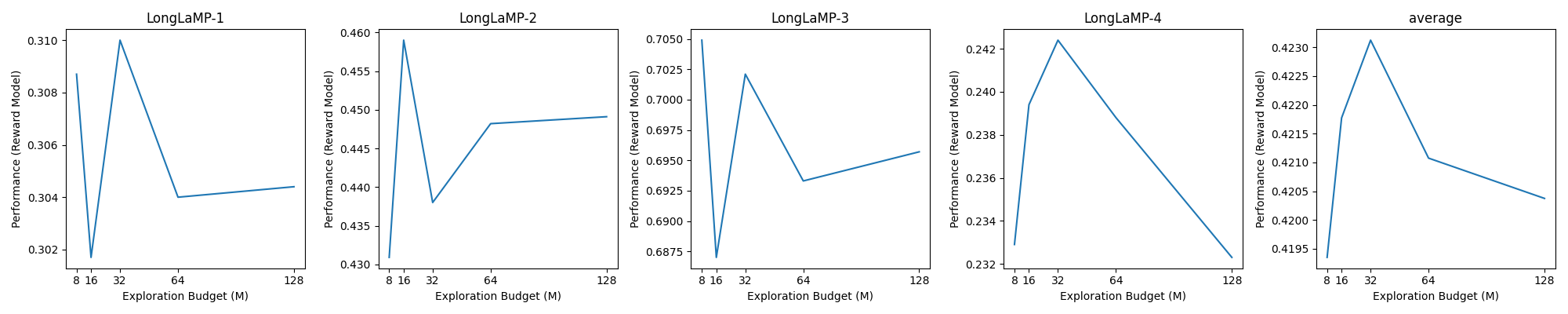}
    \vspace{-0.8cm}
    \caption{The performance of our approach with different exploration budgets ($m$) when trained for one iteration on the test set. The same plot on validation sets is depicted in Figure~\ref{fig:exploration-dev} in Appendix \ref{app:results-dev}.}
    \label{fig:exploration}
\end{figure*}

\paragraph{How does training the LLM with {REST-PG} affect the performance?}

We trained the LLM using the proposed approach, which incorporates both reasoning-enhancement and self-training. For the baselines, we evaluate LLMs that were: (1) trained using \textit{SFT} with retrieval augmentation \cite{lamp, longlamp}, (2) trained using \textit{SFT with Reasoning-Enhancement} as described in Section \ref{sec:reasoning-data-gen}, and (3) trained exclusively using self-training with \textit{ReST-EM} \cite{rest-em}. The results, shown in Table~\ref{tab:main-results}, indicate that the proposed approach, \textit{REST-PG}, outperforms all baselines across all tasks, with statistically significant improvements in 3 out of 4 tasks. Additionally, the approach shows statistically significant superior performance on average across all tasks. This demonstrates that using reasoning over personalized context, combined with self-training, can significantly enhance the performance of personalized text generation, highlighting the value of incorporating reasoning during personalized generation. The main reason for this improvement is that combining reasoning with self-training enhances the model's ability to effectively use the personalized context and align its reasoning process with the user's preferences. This, in turn, results in more tailored and accurate output for the user.

\paragraph{How does reasoning-enhancement alone affect the performance?}

To answer this question, we compare the model trained on the reasoning-enhancement data generated in Section~\ref{sec:reasoning-data-gen} and the SFT model trained on the original inputs and outputs of the LongLaMP dataset. The results of this experiment are reported in Table~\ref{tab:main-results}. These indicate that supervised fine-tuning on the generated reasoning-enhancement data from a larger model only statistically significantly improves performance on LongLaMP-3. However, there is a performance drop on the rest of the tasks, with the model performing worse than the SFT on average across all datasets, where on LongLaMP-2 this drop is statistically significant. However, on average, there is no statistically significant difference between this approach and SFT. Note that this approach underperforms compared to both methods that incorporate self-training. This observation suggests that, as discussed in our motivation, training solely on generated reasoning data is suboptimal as there is no alignment between these reasoning paths and the user's preferences for personalized text generation.

\begin{figure*}
    \centering
    \includegraphics[width=\textwidth]{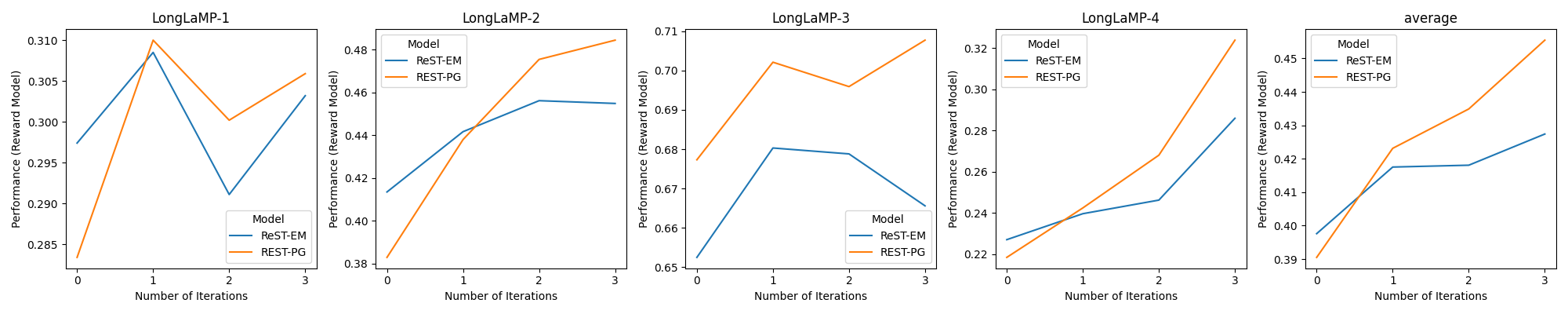}
    \caption{The effect of number of expectation-maximization steps on the performance on the test set. The same plot on validation sets is depicted in Figure~\ref{fig:iteration-performance-dev} in Appendix \ref{app:results-dev}.}
    \label{fig:iteration-performance}
    \vspace{-0.5cm}
\end{figure*}

\paragraph{How does self-training alone affects the performance?}

We trained the LLM with \textit{ReST-EM} \cite{rest-em}, similar to our approach for self-training but without considering reasoning enhancement. This approach operates similarly to ours but does not involve reasoning over the personalized context. The results of this experiment are reported in Table~\ref{tab:main-results} with the model name \textit{ReST-EM}. The results indicate that self-training significantly improves performance on LongLaMP-2 and LongLaMP-4 over both SFT and SFT with Reasoning-Enhancement. Although it improves results on LongLaMP-1 and LongLaMP-3, these improvements are not statistically significant. Moreover, it does not outperform SFT with Reasoning-Enhancement on LongLaMP-3. However, on average, this approach significantly outperforms both baselines. Note that this model is unable to outperform \textit{REST-PG} on any of the tasks, with significant differences in performance observed in 3 out of 4 tasks and in the overall average performance. This observation suggests that self-training is a promising approach for enhancing performance in personalized text generation. However, without explicitly considering the user's implicit preferences or writing style, the improvement on personalized text generation tasks is limited.

\paragraph{How does the exploration budget affect the performance of \textit{REST-PG}?}

We apply our method using different exploration budgets $m$ during the expectation step, generating 8, 16, 32, 64, and 128 outputs per input and train the LLM for one iteration on them. The results are shown in Figure~\ref{fig:exploration}. While different tasks benefit from varying exploration budgets, on average, increasing this exploration budget improves the results up to a certain point before decreasing the performance. This suggests that overly increasing the exploration budget may not be beneficial; as the model generates more examples, the diversity among high-reward examples can negatively impact the model's performance. Therefore, tuning this parameter considerably affects performance.

\paragraph{How does the number of training iterations affect the performance?}

We vary the number of training iterations for self-training models, \textit{ReST-EM} and \textit{REST-PG}, and evaluate them after each iteration. The results are illustrated in Figure~\ref{fig:iteration-performance}. This figure suggests that, on average, increasing the number of iterations leads to improvements in both models. However, the performance gap between the models widens as the number of iterations increases, with \textit{REST-PG} consistently outperforming \textit{ReST-EM}. Additionally, while without any self-training, the \textit{SFT with reasoning-enhancement} performs worse than the \textit{SFT} on 3 out of 4 tasks, after just one iteration of self-training, \textit{REST-PG} surpasses \textit{ReST-EM} in all tasks. This show that while both benefit from more iterations, improvements are more pronounced for \textit{REST-PG}, as additional iterations allow the model to discover more effective reasoning paths, further enhancing its performance.

\paragraph{Is it better to start from a base checkpoint or continue training from the SFT?}

\begin{figure}
    \centering
    \includegraphics[width=0.9\linewidth]{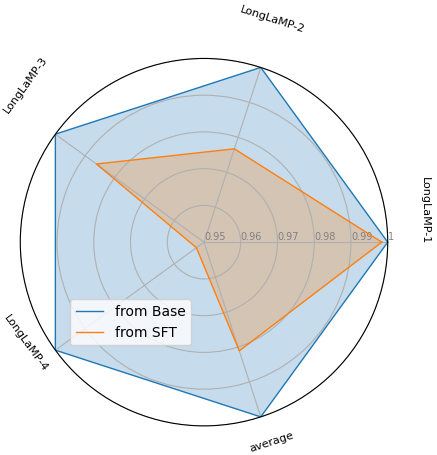}
    \vspace{-0.2cm}
    \caption{The relative performance of our model trained from the base checkpoint and the SFT checkpoint for one iteration on the test set. The same plot on validation sets is depicted in Figure~\ref{fig:strat-sft-base-dev} in Appendix \ref{app:results-dev}.}
    \label{fig:strat-sft-base}
    \vspace{-0.5cm}
\end{figure}

We train two models using the proposed approach: one starting from the base checkpoint and the other from the SFT checkpoint, which was trained with the data generated in Section \ref{sec:reasoning-data-gen}. We plotted the relative performance of these two models after training using our approach in Figure~\ref{fig:strat-sft-base}. This figure demonstrates that the model starting from the SFT checkpoint underperforms compared to the model starting from the base checkpoint across all tasks, achieving only 96\% of the performance of the latter on average. This suggests that starting from a new base model in each iteration is more effective. We believe this is because starting from a base checkpoint allows the model to learn reasoning paths more freely, without being constrained by patterns learned during previous training iteration.

\subsection{Case Study}

To compare the generated outputs using our approach, we provide two categories of examples.

\paragraph{Improvements in the final generated response.}

Figure~\ref{fig:case-study-abstract} in Appendix~\ref{app:case-study} shows an output generated by \textit{REST-PG} and \textit{ReST-EM} for a prompt from the personalized abstract generation dataset. \textit{REST-PG} provides a more precise description of the proposed method and correctly predicts the evaluation dataset, ImageNet, while \textit{ReST-EM} produces a hallucinated and incorrect prediction. This example highlights that \textit{REST-PG} better utilizes the user's personalized context to generate more accurate and personalized response. In this case, \textit{REST-PG}'s correct prediction was guided by the author's previous experiments on the ImageNet dataset.

\paragraph{Improvements in reasoning path toward generating the final response.}

Figure \ref{fig:case-study-review} in Appendix \ref{app:case-study} shows an example of personalized output generated by \textit{REST-PG} and \textit{SFT with Reasoning-Enhancement} for a prompt from the personalized review writing dataset. Here, \textit{SFT with Reasoning-Enhancement} introduced some hallucinated names in the reasoning, which were carried over into the final output. In contrast, \textit{REST-PG} successfully avoided this issue by recognizing that adding inaccurate details negatively affects the reward model's evaluation. Notably, \textit{REST-PG} inferred that the user ``\textit{values well-developed characters and relationships}'' and incorporated this into the review, aligning closely with the expected output.

\section{Related Work}

\paragraph{Personalization}

is an important topic with use cases in search, recommendation, and text generation \cite{10.1145/2702123.2702503, 10.1145/1462198.1462203, naumov2019deep, lamp}. \citet{lamp} introduced a Retrieval-Augmented Generation (RAG)-based method for personalizing LLMs and the LaMP benchmark for evaluating short-form personalized text generation. \citet{longlamp} extended this by introducing the LongLaMP benchmark for long-form personalized text generation. Another direction has focused on designing personalized writing assistants \cite{li2023teach, mysore2023pearl, lu2024corporate} and agents \cite{zhang-etal-2024-llm-based}. Efforts to personalize LLMs include training retrieval models based on feedback for text generation \cite{rspg}, optimizing LLMs with personalized feedback \cite{jang2023personalized}, and automatic personalized prompt generation \cite{Li_2024}. Recent studies have explored parameter-efficient fine-tuning \cite{tan2024personalized} and their integration with RAG \cite{peft-rag-personalization}. This paper differs itself by focusing on training LLMs to effectively leverage personalized context and incorporate reasoning into output generation.

\paragraph{Reasoning-Enhancement in LLMs}

is the model's ability to think step-by-step, also known as chain-of-thoughts (CoT), before responding to prompts. This improves performance of LLMs in complex tasks such as mathematical, logical, and commonsense reasoning \cite{cot,liu2023logicot, 10.1007/978-981-97-7232-2_13}. Additionally, smaller LLMs can acquire this ability through distillation from larger models \cite{li-etal-2023-symbolic}. Reasoning-enhancement has not been studied for personalization due to difficulty of understanding user's implicit intent and collecting data to train LLMs for this ability. This paper focuses on training LLMs to achieve this using RL. Concurrently, OpenAI released O1 \cite{openai_o1}, incorporating reasoning into response generation, focusing on math and logical problems.

\paragraph{Self-Training}

is a new paradigm in which LLMs generate the training data for themselves \cite{amini2024selftrainingsurvey}. Here, the LLM generates outputs for given inputs, and those that are of high quality, assessed by a reward function, are used to train the model further \cite{rest-em, star}. \citet{rest-em} employ expectation maximization with RL to optimize the model on self-generated outputs, focusing on math and code generation. Similarly, \citet{star} use CoT prompting to generate answers for math and commonsense problems, utilizing only those that lead to correct answers for training the model. Extensions to both approaches includes improved rewarding mechanism \cite{rest-tree} and generating per token rationals. Our work differs from prior studies in key aspects. Previous work focuses on math reasoning and code generation, where multiple-choice or clearly defined correct answers are present. Conversely, free-form personalized generation lacks a definitive correct or incorrect answer because an output might be desirable for one user but not for the others. Additionally, our approach extends the work of \citet{rest-em} by incorporating reasoning into response generation, allowing for more personalized responses.

\section{Conclusions}

This paper proposes REST-PG, a multi-stage framework designed to train LLMs to reason over personalized contexts during response generation. The framework begins by instructing the LLM to generate a reasoning path, based on the input, expected output, and personalized context, outlining how the final output should be derived. This reasoning paths are then used to train the LLM to generate both the reasoning steps and the response in a single inference path, instilling a preliminary reasoning ability in the LLM. Following this, we apply expectation-maximization reinforced self-training to iteratively align the model's reasoning with the user's preferences based on a reward function that evaluates the similarity between the generated response and the expected output for the user. Our results on the LongLaMP benchmark show that our approach significantly outperforms supervised fine-tuning, achieving 14.5\% improvement, and it outperforms self-training without reasoning by 6.5\% in personalized text generation. Additionally, we conduct a detailed ablation study which provides insights into various aspects of our proposed method.

\section*{Limitations}

This work has limitations concerning both evaluation and latency of the proposed approaches.

\paragraph{Evaluation of Long-Form Personalized Text Generation.}

Evaluating personalization in text generation presents inherent challenges, as the ideal judge for the outputs would be the individual who created the inputs \cite{lamp}. Unfortunately, accessing these original users for existing datasets is often unfeasible. Furthermore, human evaluation remains difficult, as it’s not guaranteed that annotators can accurately assess whether the output meets the original prompt writer's expectations. Additionally, as highlighted in previous studies, evaluating long-form text generation is a complex and active area of research in the natural language processing community \cite{koh-etal-2022-far, krishna-etal-2021-hurdles, belz-reiter-2006-comparing}. In this paper, we combine these two challenging concepts, which further complicates the evaluation process.

To the best of our knowledge, there is currently no widely accepted metric for evaluating generated personalized outputs. Traditional metrics, such as ROUGE \cite{rouge} and BLEU \cite{bleu}, which rely on term matching, have proven inadequate for assessing long-form text generation \cite{koh-etal-2022-far, krishna-etal-2021-hurdles, belz-reiter-2006-comparing}. Recent efforts in the community have shifted toward utilizing LLMs as evaluators \cite{li2024leveraginglargelanguagemodels}. Given that we have access to the expected output for each user, we follow the same approach and employ LLMs to assess the similarity between the generated output and the expected output for that specific user. While this evaluation method is not perfect, it represents the most effective approach available within the constraints.

\paragraph{Latency of Reasoning During Response Generation.}

While incorporating reasoning over personalized context in this paper leads to substantial improvements in the quality of the final generated output, it also introduces a trade-off: an increase in the overall output length. This extended length, when processed by a standard transformer-based LLM, results in a rise in decoding time. This study, however, does not address or attempt to optimize this increased decoding overhead by reasoning-enhancement. While the current focus is on enhancing output quality and personalization, future research could explore strategies to mitigate these computational costs.


\bibliography{custom}

\appendix

\section{Large Language Model Evaluator \& Human Evaluation}
\label{app:human-eval}

Although the LongLaMP benchmark \cite{longlamp} primarily relies on ROUGE \cite{rouge} to assess the quality of long-form text generation, prior studies suggest that lexical overlap metrics often fail to capture semantic similarities \cite{bert-score}, especially in long-form generation tasks \cite{koh-etal-2022-far, krishna-etal-2021-hurdles, belz-reiter-2006-comparing}. Following the approach proposed by \citet{geval}, we employ an instruction-tuned LLM, Gemma \cite{gemma}, with 7 billion parameters as our text similarity evaluator. Since this LLM is trained on large instruction tuning datasets, if provided with a well-defined evaluation instruction, it can serve as effective judges for text similarity tasks \cite{li2024leveraginglargelanguagemodels}. 

\begin{figure}[!ht]
    \centering
    \includegraphics[width=\linewidth]{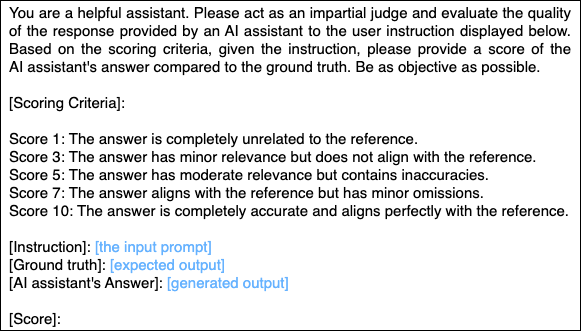}
    \caption{The prompt used for reward model to evaluate the generated text based on the input, the reference output, and the provided criteria.}
    \label{fig:eval-prompt}
\end{figure}

Following \citet{geval}, to evaluate the generated outputs, we feed the evaluator LLM with the input prompt, the generated text, and the reference output, accompanied by a prompt that explains the evaluation criteria (as depicted in Figure \ref{fig:eval-prompt}). In this prompt, the criteria that determine whether the generated output receives the defined score are clearly outlined. After feeding the model with the prompt (containing the input, expected output, and generated output), the LLM evaluator calculates the probability for each score defined by the criteria in the prompt. The score with the highest probability is selected as the score for the generated output for this input. To normalize the score and ensure it falls within the range of 0 to 1, the selected score is divided by 10 (i.e., the maximum score that the LLM evaluator can assign to an output). This normalized score reflects the model's assessment of the generated output based on the predefined criteria. To validate whether the LLM evaluator model can accurately assess the quality of generated texts, we design two experiments. 

In the first experiment, we conducted a human evaluation to validate the LLM evaluator. Annotators were presented with 100 pairs of generated texts from the models discussed in this paper. For each pair, the annotators were asked to select the text that best reflected the expected output given the input. The pairs were selected such that there was a score difference of at least 0.5 between the two texts, as determined by our LLM evaluator model. The results of the human evaluation indicate that our metric aligns with human judgment in 73\% of the cases. Additionally, the metric shows a correlation of 0.46 with human judgment, suggesting that the LLM evaluator model generally agrees with human assessments. Note that previous studies on designing automatic metrics for personalized text generation have highlighted that such approaches may struggle to achieve very high agreement with human evaluations. This is because personalized text generation is inherently subjective, and only the individual who wrote the input can fully assess whether the generated output meets their expectations or preferences \cite{wang2023automatedevaluationpersonalizedtext}. Since access to these specific annotators is not possible for existing datasets, this type of evaluation may not provide a completely reliable measure of the quality of personalized text generation.

\begin{figure*}
    \centering
    \includegraphics[width=\textwidth]{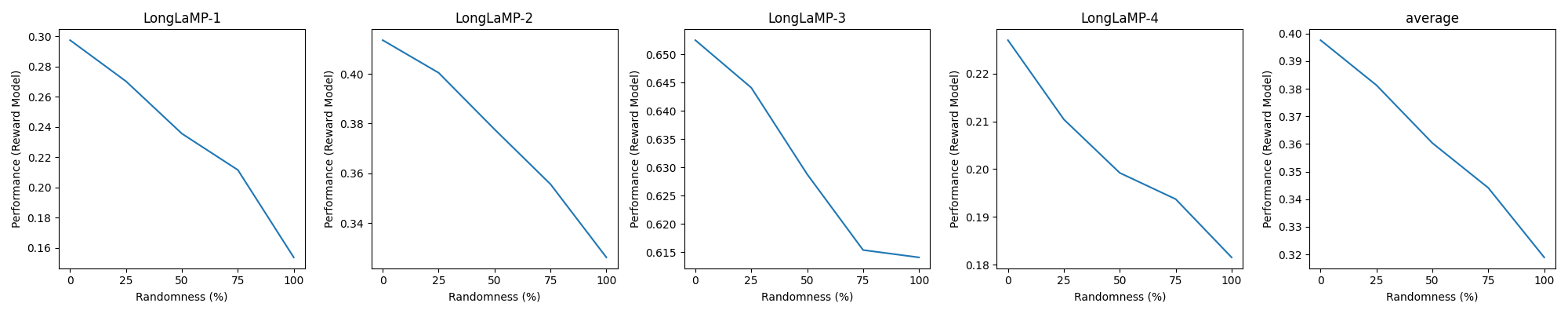}
    \caption{The affect of randomly shuffling profiles on the reward model's scores.}
    \label{fig:reward-random}
\end{figure*}

To further evaluate the LLM evaluator, we designed an experiment in which the model trained on the LongLaMP benchmark using supervised fine-tuning (as detailed in Section \ref{sec:experiments}) is tested with personalized contexts that are randomly assigned to inputs at varying rates. Specifically, we randomly replaced $S$ percent of the personalized contexts with those from other users, while keeping the input prompt and expected output unchanged. This experiment aims to determine whether the LLM evaluator can detect changes in the personalized context based on the generated text and its comparison with the expected output. The results of this experiment are shown in Figure \ref{fig:reward-random}. The figure illustrates that as the rate of random sampling increases, the LLM evaluator linearly assigns lower scores to the texts generated by the same model. This suggests that the LLM evaluator is linearly sensitive to discrepancies in the generated text context from unmatched personalized context with the expected output for the given input. 

Therefore, considering both experiments, we believe and are convinced that the LLM evaluator used in this paper is capable of evaluating the quality of generated personalized text when a personalized expected output is provided. These findings demonstrate that the LLM evaluator can effectively align with human judgments and is sensitive to changes in personalized context, supporting its utility for assessing personalized text generation.

\section{Detailed Experiments Setup}
\label{app:setup}

This section outlines the detailed configuration of the experiments conducted in this paper.

\paragraph{Datasets \& Tasks.}

In this paper, we utilize the LongLaMP benchmark \cite{longlamp} to conduct our experiments, which consists of four personalized long-form text generation tasks: 

\begin{enumerate}
    \item Personalized Email Completion: Given an input email, the task is to generate a personalized continuation based on the user's writing style and preferences.
    
    \item Personalized Abstract Generation: This task involves generating personalized abstracts for technical documents or articles given the title and some keywords, reflecting the user's writing patterns and focus areas.
    
    \item Personalized Review Writing: The model generates personalized product reviews that reflect the user's preferences, given the description of the product and the score that is assigned to the product by the user.
    
    \item Personalized Topic Writing: For a post summary on a topic, the task is to generate a personalized long-form full post that reflects the user’s writing style, preferences, and opinion on topic.
\end{enumerate}

Each example in the dataset represents a distinct user and includes, an input prompt relevant to the task, an expected output tailored to that specific user, and a user profile containing historical data, such as previously generated texts, to capture the user’s writing habits and preferences. We utilize the user-based setting of the LongLaMP benchmark to perform our experiments. The statistics of the datasets are reported in Table~\ref{tab:task-stats}.

\paragraph{Training Setup.}

We utilize the Gemma model \cite{gemma} with 2 billion parameters as the LLM. To incorporate personalized context, we follow the retrieval-augmented generation approach for personalized text generation, as described in \citet{lamp}, with the prompt shown in Figure~\ref{fig:cot-input-output-template}  in Appendix \ref{app:prompts}. We employ multi-task learning to train a single model across all tasks in the LongLaMP benchmark, allowing the model to generalize and perform well on diverse personalized text generation tasks. We retrieve $k = 5$ items from the user profile using Contriever \cite{contriever}. Following \citet{rest-em}, the models are trained over $T = 3$ iterations, generating $m = 32$ outputs per input during the expectation step, with a decoding temperature of $\gamma = 0.7$ using Nucleus Sampling \cite{Holtzman2020The}, unless otherwise specified. We set the output selection threshold to $\tau = 1.0$, and at each iteration, the training begins from a new, untrained checkpoint unless otherwise stated.

For each iteration of training, we use the Adafactor optimizer \cite{adafactor} with a learning rate of $5 \times 10^{-6}$ and a linear learning rate decay of $0.1$, along with $250$ warmup steps, for a maximum of 10,000 training steps. The batch size is set to 64, and we apply a weight decay of $0.01$. We also utilize a gradient cliping of $1.0$ for optimization. The input length is limited to a maximum of 5,120 tokens, and the output is capped at 1,536 tokens. The experiments are conducted on 64 TPU-v4 \cite{tpuv4} cores, each with 32GB of memory, for a maximum duration of 1 day. All reported results are based on a single run.

\paragraph{Inference Setup.}

During inference, we limit the input to a maximum of 5,120 tokens and the output to 1536 tokens, where we use nucleus sampling \cite{Holtzman2020The} with a sampling temperature of $\gamma = 0.1$ to produce more deterministic outputs from the LLM. For evaluation, models are assessed using full precision on the entire test dataset. However, during checkpoint validation in the training phase, we randomly sample 1,024 examples from the validation set to evaluate the model and choose the best checkpoint every 1000 steps. Inference is conducted on the same infrastructure and resources used during the training setup.

\begin{table*}[!ht]
    \centering
    \begin{adjustbox}{max width=\textwidth}    
        \begin{tabular}{l|cccccc}
        \toprule
            \textbf{Task} & \textbf{\#train} & \textbf{\#validation} & \textbf{\#test} & \textbf{Input Length} & \textbf{Output Length} & \textbf{Profile Size} \\
            
            \midrule
            {LongLaMP-1: Personalized Email Completion} & 3286 & 958 & 823 & 46.45 $\pm$ 21.45 & 92.59 $\pm$ 60.68 & 85.65 $\pm$ 51.67 \\
            
            \midrule
            {LongLaMP-2: Personalized Abstract Generation} & 13693 & 4560 & 4560 & 33.82 $\pm$ 5.71 & 144.28 $\pm$ 68.40 & 120.30 $\pm$ 118.81  \\
            
            \midrule
            {LongLaMP-3: Personalized Review Writing} & 14745 & 1826 & 1822 & 119.39 $\pm$ 73.06 & 304.54 $\pm$ 228.61 & 34.39 $\pm$ 57.31  \\
            
            \midrule
            {LongLaMP-4: Personalized Topic Writing} & 11442 & 2452 & 2453 & 28.36 $\pm$ 36.08 & 263.03 $\pm$ 243.34 & 50.39 $\pm$ 2898.60  \\
        \bottomrule
        \end{tabular}
    \end{adjustbox}
    \caption{The statistics of the datasets in the LongLaMP benchmark on user-based setting.}
    \label{tab:task-stats}
\end{table*}

\section{Overview of Prompts and Templates}
\label{app:prompts}

\begin{figure*}
    \centering
    \includegraphics[width=\linewidth]{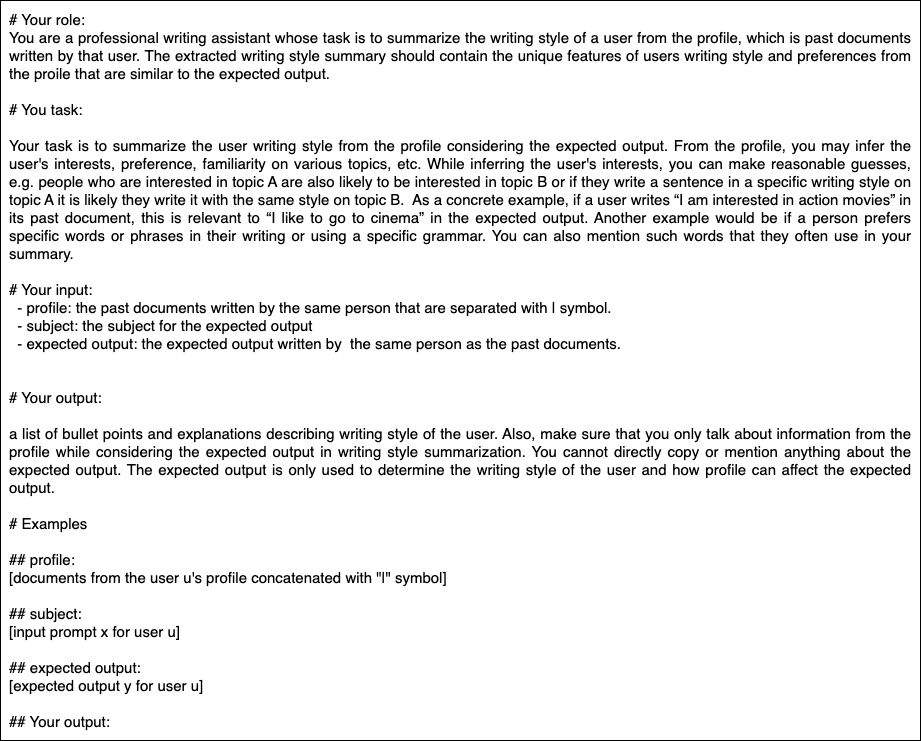}
    \caption{The prompt used to generate summary of user's preferences, interests, background knowledge, and writing style features as a reasoning method over the personalized context.}
    \label{fig:summary_gen_prompt}
\end{figure*}

We utilize an instruction-tuned Gemma \cite{gemma} LLM with 7 billion parameters to generate the initial reasoning over personalized context data. These reasoning data is used to train the model to develop a preliminary reasoning ability over personalized user context. The prompt used to generate such data is shown in Figure~\ref{fig:summary_gen_prompt}. This prompt encourages the model to consider both the final output and the input prompt, and based on this, generate a summary of user's preferences, interests, background knowledge, and writing style features from the user's personalized context that reflects their interests, preferences, and familiarity with various topics. Additionally, the prompt encourages the model to make reasonable inferences about the user’s preferences on different topics. For instance, if a user writes in a specific writing style on a particular topic, the model may infer that the user is likely to use a similar approach for other related topics as well.

\begin{figure*}
    \centering
    \includegraphics[width=\linewidth]{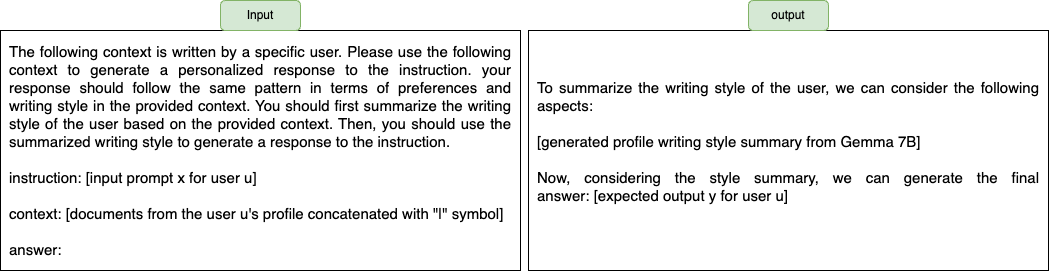}
    \caption{The input output template used for training the model with reasoning-enhancement data.}
    \label{fig:cot-input-output-template}
\end{figure*}

Finally, to train the LLM to perform reasoning over personalized context during output generation, the generated reasoning data is combined with the expected output using a predefined template, as shown in Figure~\ref{fig:cot-input-output-template}. This template enables the model to generate personalized responses by incorporating reasoning based on the user’s preferences. The model is fed with an input consisting of the user’s prompt and personalized context. The model is first tasked with generating a summary of user's preferences and writing style features based on the input, which is then followed by generating the final response to the prompt. The combined output—--both the reasoning path and the final response—--serves as the expected output in this template. Essentially, the model is trained to generate both the reasoning steps and the final response in a single inference pass.

\begin{table*}[!ht]
    \centering
    \adjustbox{max width=\textwidth}{
    \begin{tabular}{ll|c|c|c|c||c}
        \toprule
        &\multirow{3}{*}{\textbf{Model}} & \textbf{LongLaMP-1:} & \textbf{LongLaMP-2:} & \textbf{LongLaMP-3:} & \textbf{LongLaMP-4:} & \multirow{2}{*}{\textbf{Average}} \\
        & & \textbf{Personalized} & \textbf{Personalized} & \textbf{Personalized} & \textbf{Personalized} & \\
        & & \textbf{Email Completion} & \textbf{Abstract Generation} & \textbf{Review Writing} & \textbf{Topic Writing} & \textbf{(macro)} \\
         \midrule
        1 & SFT  & 0.3672 & 0.4046 & 0.6455 & 0.2293 & 0.4116 \\
        
        2 & SFT w/ Reasoning-Enhancement & 0.3426 & 0.3824 & 0.7181 & 0.2495 & 0.4231 \\

        3 & ReST-EM  & 0.3711 & 0.4550 & 0.6664 & 0.2853 & 0.4444 \\
        \midrule
        
        & \textbf{REST-PG} & \textbf{0.3800} & \textbf{0.4827} & \textbf{0.7197} & \textbf{0.3561} & \textbf{0.4846} \\
        \bottomrule
    \end{tabular}}
    \caption{The performance of all methods on the validation sets of the LongLaMP benchmark. In order to speed up the experiments, a maximum of 1,024 samples from each task randomly was selected, instead of evaluating on the full validation set.}
    \label{tab:main-results-dev}
\end{table*}

\begin{figure*}
    \centering
    \includegraphics[width=\textwidth]{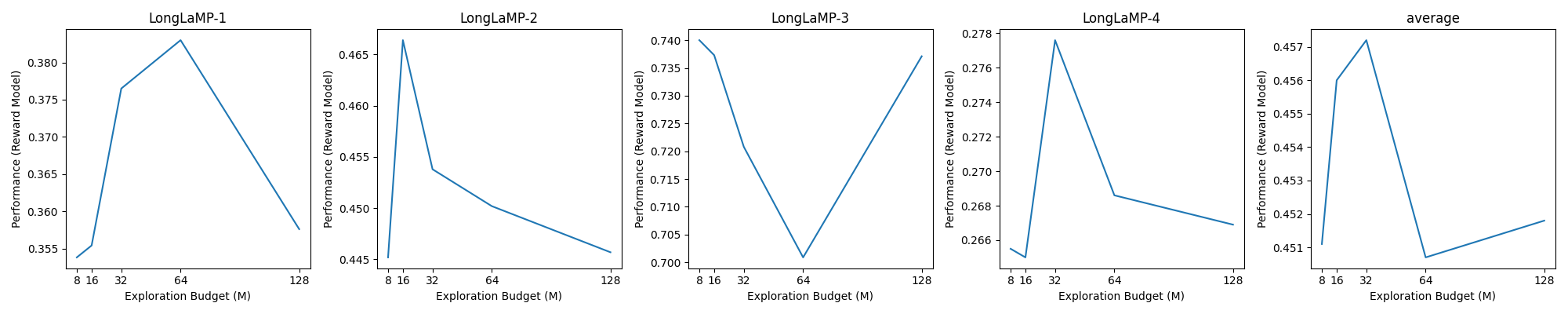}
    \caption{The performance of our approach with different exploration budgets ($m$) when trained for one iteration on the validation set. In order to speed up the experiments, a maximum of 1,024 samples from each task randomly was selected, instead of evaluating on the full validation set.}
    \label{fig:exploration-dev}
\end{figure*}

\section{Results on the Validation Sets}
\label{app:results-dev}

This section reports the results of the experiments performed in Section~\ref{sec:experiments} on the validation set of the datasets in the LongLaMP benchmark \cite{longlamp}. To accelerate the training phase, we randomly selected at most 1,024 samples from each dataset and evaluated the checkpoints on those samples. Therefore, the results presented here are not based on the entire validation set of the datasets.

The results of baselines and the proposed approaches on the dev set are reported in Table~\ref{tab:main-results-dev}. The results in this table suggest that SFT with reasoning-enhancement, unlike on the test set, was able to help the model outperform the SFT model without reasoning. Additionally, we observe that self-training without reasoning using ReST-EM outperforms the SFT baselines, similar to the results on the test set. Finally, REST-PG outperforms all the baselines across all tasks in the LongLaMP benchmark, consistent with the test set results.

The results of the experiment on varying the exploration budget in the expectation step of self-training on the dev set are shown in Figure~\ref{fig:exploration-dev}. Similar to the test set, the results indicate that while different tasks may benefit from different budgets, on average, generating 32 outputs leads to the best performance. This again emphasizes the importance of hyper-parameter tuning for this approach.

\begin{figure*}
    \centering
    \includegraphics[width=\textwidth]{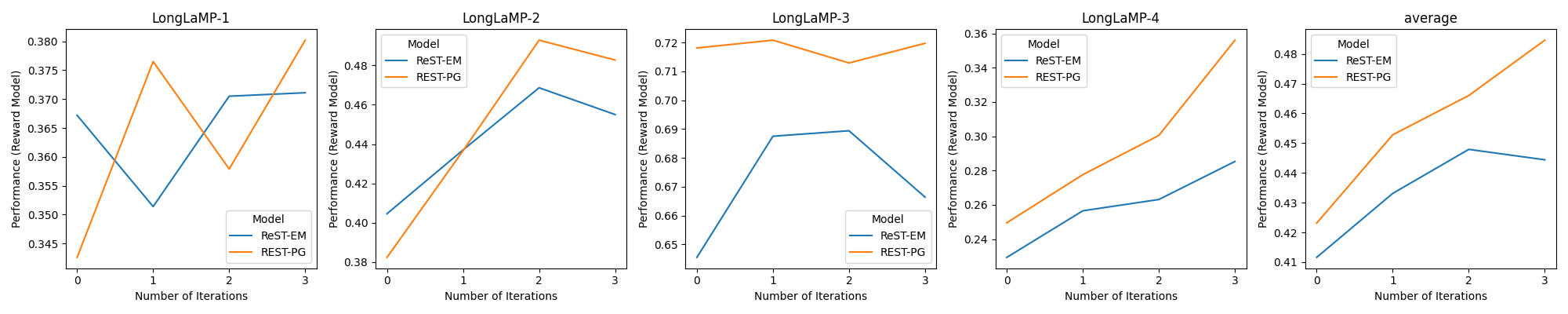}
    \caption{The affect of number of expectation-maximization steps on the performance on the validation set. In order to speed up the experiments, a maximum of 1,024 samples from each task randomly was selected, instead of evaluating on the full validation set.}
    \label{fig:iteration-performance-dev}
\end{figure*}

The results of the experiments on varying the number of training iterations are reported in Figure~\ref{fig:iteration-performance-dev}. This figure suggests that, similar to the test set, increasing the number of iterations leads to improved performance for both ReST-EM and REST-PG. The gap between their performance grows as iterations increase, showing that REST-PG benefits more from additional iterations. Note that after just one iteration, REST-PG outperforms ReST-EM on all datasets, even on those that performed worse with reasoning-enhancement before self-training.

\begin{figure}[!ht]
    \centering
    \includegraphics[width=0.8\linewidth]{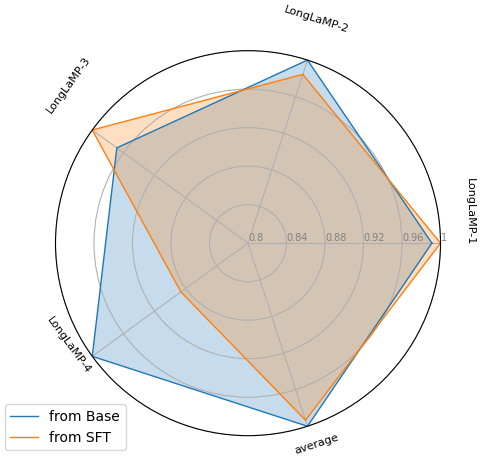}
    \caption{The relative performance of our model trained from the base checkpoint and the SFT checkpoint for one iteration on the validation set. In order to speed up the experiments, a maximum of 1,024 samples from each task randomly was selected, instead of evaluating on the full validation set.}
    \label{fig:strat-sft-base-dev}
\end{figure}

Finally, the results of experiments on starting from a new base checkpoint or continuing training from the previous checkpoint are reported in Figure~\ref{fig:strat-sft-base-dev}. Similar to the test set, the results show that, on average, starting from a fresh base checkpoint performs better than continuing training from the previous checkpoint. This finding reinforces the idea that initializing from a fresh checkpoint leads to improved performance compared to fine-tuning from previously trained models.

\section{Case Study \& Output Examples}
\label{app:case-study}

This section presents samples of the outputs generated at various stages of our approach.

\paragraph{Generated reasoning path using Gemma 7B given input, output, and personalized context.}

\begin{figure*}
    \centering
    \includegraphics[width=\textwidth]{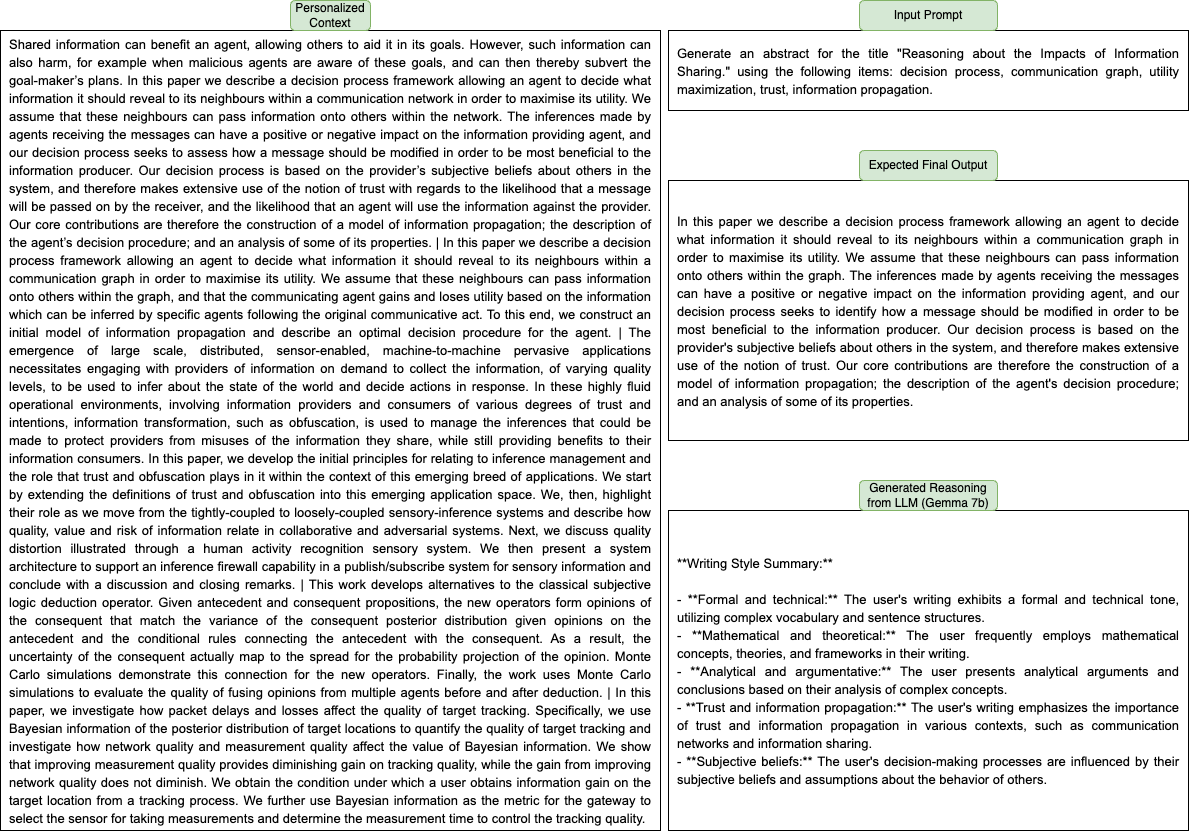}
    \caption{The generated profile summary with Gemma 7B on personalized abstract generation task.}
    \label{fig:generated-summary-abstract}
\end{figure*}

\begin{figure*}
    \centering
    \includegraphics[width=\textwidth]{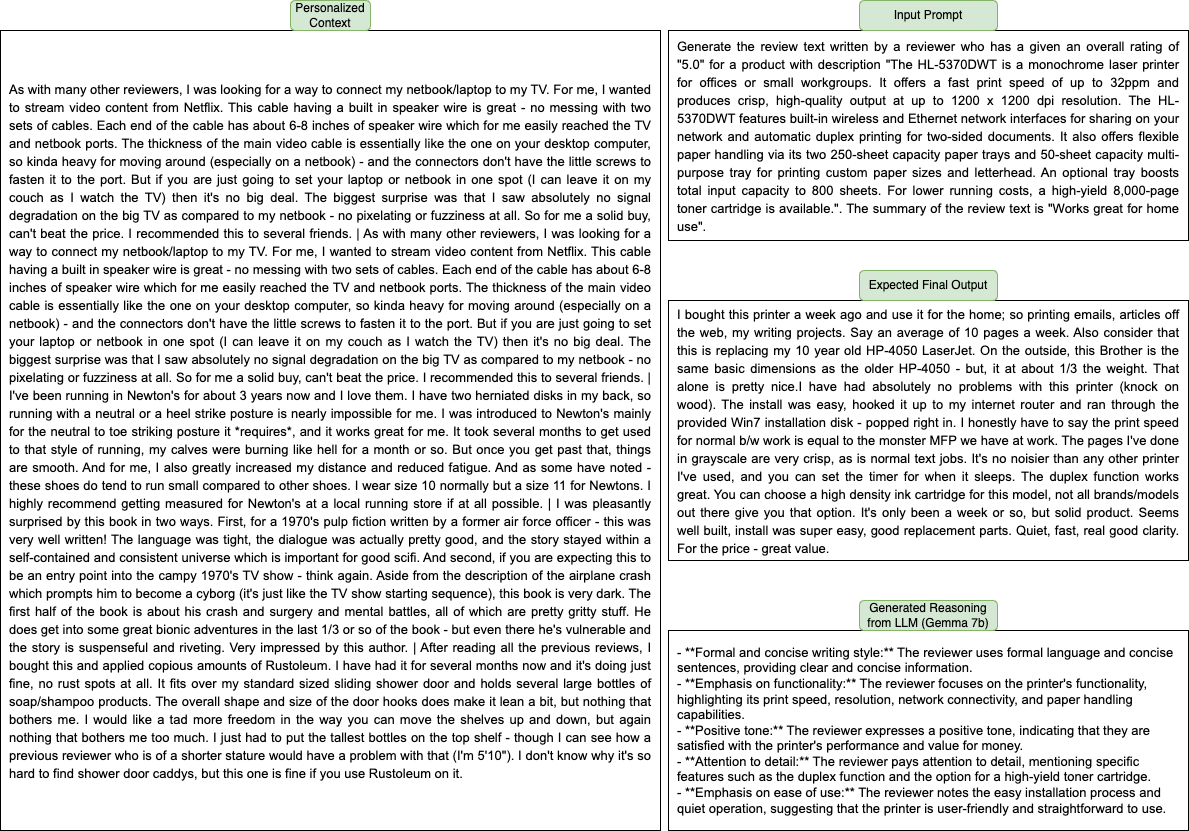}
    \caption{The generated profile summary with Gemma 7B on personalized review writing task.}
    \label{fig:generated-summary-review}
\end{figure*}

As explained in Section~\ref{sec:reasoning-data-gen}, we utilize the Gemma 7B model to generate reasoning over personalized context by considering the personalized context, input prompt, and expected output. Figures \ref{fig:generated-summary-abstract} and \ref{fig:generated-summary-review} showcase two examples of such reasoning outputs. These generated reasoning summaries are subsequently used to train a smaller model, enabling it to develop preliminary reasoning abilities during the generation of responses.

\paragraph{Improvements in the final generated response.}

\begin{figure*}
    \centering
    \includegraphics[width=\textwidth]{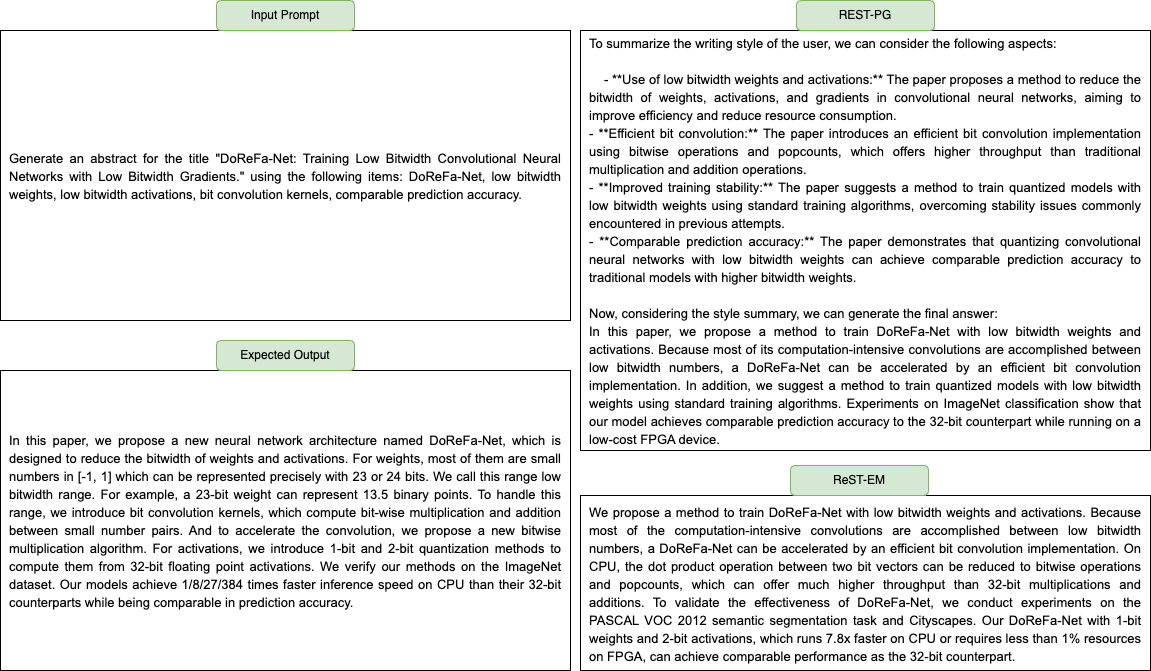}
    \caption{The comparison between answer generated with REST-PG and ReST-EM on an example from personalized abstract generation task.}
    \label{fig:case-study-abstract}
\end{figure*}

Figure~\ref{fig:case-study-abstract} provides an example of personalized output generated by REST-PG and ReST-EM for a prompt from the personalized abstract generation dataset. The REST-PG model delivers a more accurate description of the proposed method and correctly predicts the evaluation dataset, ImageNet, while the ReST-EM model hallucinates and provides an incorrect guess. This example illustrates that REST-PG more effectively leverages the user's past history to generate more accurate and personalized text. In this case, the author's previous experiments on the ImageNet dataset helped the model make the correct prediction.

\paragraph{Improvements in reasoning path toward generating the final response.}

\begin{figure*}
    \centering
    \includegraphics[width=\textwidth]{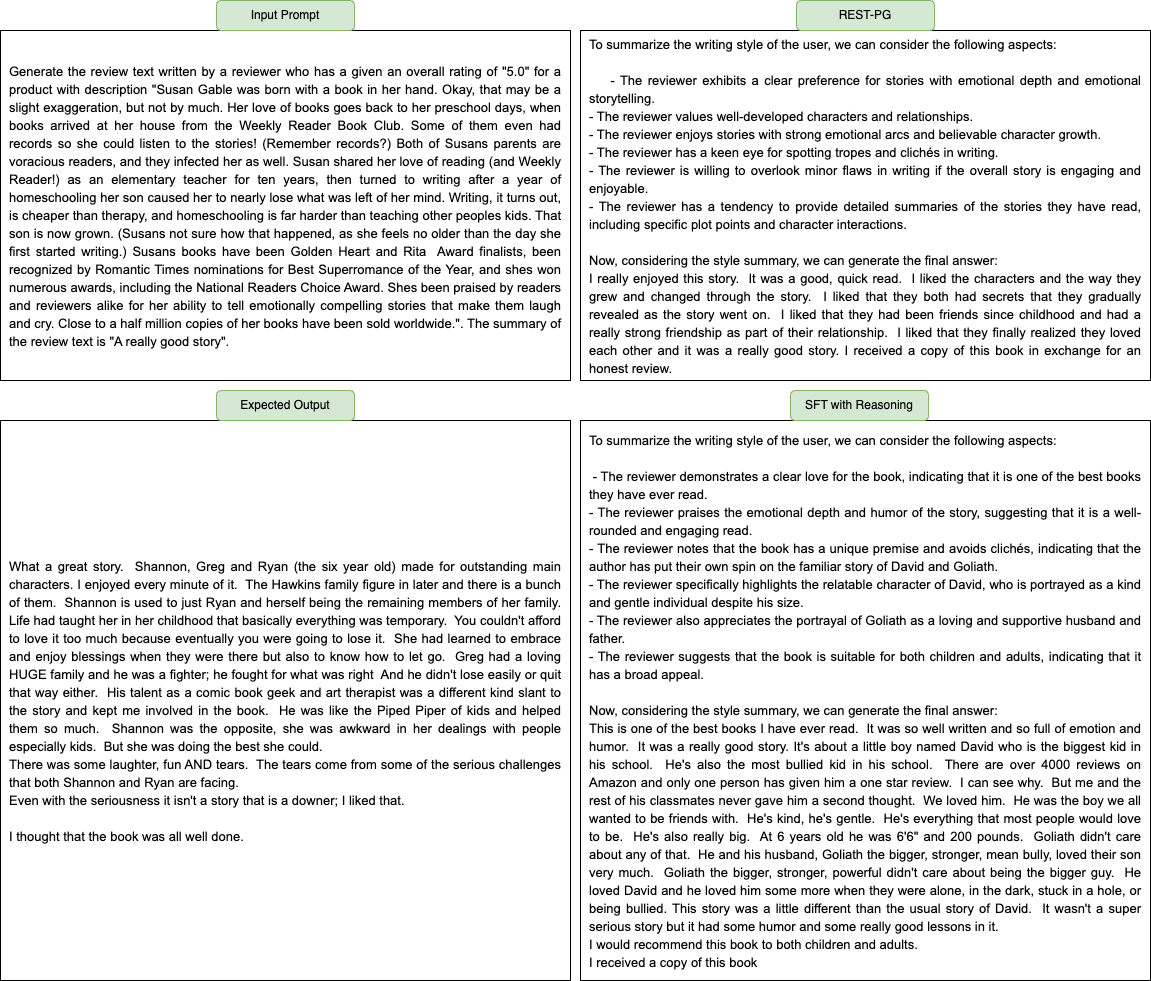}
    \caption{The comparison between answer generated with REST-PG and SFT with Reasoning-Enhancement on an example from personalized review writing task.}
    \label{fig:case-study-review}
\end{figure*}

Figure~\ref{fig:case-study-review} shows an example of outputs generated by REST-PG and SFT with Reasoning-Enhancement for a given prompt from the personalized review writing dataset. In this case, the SFT with Reasoning-Enhancement model hallucinated some names in the reasoning path and incorporated them into the final generated output. In contrast, REST-PG effectively avoided such reasoning, as it recognizes that introducing inaccurate details negatively impacts the reward model's assessment. Additionally, an interesting observation is that REST-PG inferred that the user \textit{"values well-developed characters and relationships"} and reflected this in the review text, aligning with the expected output.

\end{document}